# AUTOMATED DRIVING ARCHITECTURE AND OPERATION OF A LIGHT COMMERCIAL VEHICLE

**Murat Gözü[1], Mümin Tolga Emirler[2*], İsmail Meriç Can Uygan[3],**
**Tevfik Ali Böke[4], Levent Güvenç[5,6,7], Bilin Aksun Güvenç[5,6]**

[1] İstanbul Okan University, Institute of Sciences and Engineering, Mechatronics Engineering Program, İstanbul, Turkey
[2] Yıldız Technical University, Faculty of Applied Sciences, Department of Aviation Electronics, İstanbul, Turkey
[3] TÜBİTAK BİLGEM Cyber Security Institute, Gebze, Kocaeli, Turkey
[4] TOFAŞ Research and Development Center, Bursa, Turkey
[5] The Ohio State University, Automated Driving Lab, Columbus, Ohio, USA
[6] The Ohio State University, Department of Mechanical and Aerospace Engineering, Columbus, Ohio, USA
[7] The Ohio State University, Department of Electrical and Computer Engineering, Columbus, Ohio, USA
[*] Corresponding Author: Mümin Tolga Emirler, tolgaemirler@yahoo.com

**ABSTRACT**

This paper is on the automated driving architecture and operation of a light commercial vehicle. Simple longitudinal and lateral dynamic models of the vehicle and a more detailed CarSim model are developed and used in simulations and controller design and evaluation. Experimental validation is used to make sure that the models used represent the actual response of the vehicle as closely as possible. The vehicle is made drive-by-wire by interfacing with the existing throttle-by-wire, by adding an active vacuum booster for brake-by-wire and by adding a steering actuator for steer-by-wire operation. Vehicle localization is achieved by using a GPS sensor integrated with six axes IMU with a built-in INS algorithm and a digital compass for heading information. Front looking radar, lidar and camera are used for environmental sensing. Communication with the road infrastructure and other vehicles is made possible by a vehicle to vehicle communication modem. A dedicated computer under real time Linux is used to collect, process and distribute sensor information. A dSPACE MicroAutoBox is used for drive-by-wire controls. CACC based longitudinal control and path tracking of a map of GPS waypoints are used to present the operation of this automated driving vehicle.

## 1. INTRODUCTION

Connected and automated driving has become an important area of research for automotive OEMs as series production of connected and automated driving vehicles is expected to start after 2020 [1]. Connected and automated driving work has concentrated mostly on passenger cars [2-5]. However, the largest impact of these new technologies is expected to be for commercial vehicles. The light commercial vehicle application is unique and their drivers would benefit most from automated driving since drivers of these vehicles have to drive for long amounts of time every day. Consequently, this paper concentrates on the use of a light commercial vehicle as an automated driving vehicle platform.

This paper focuses on the technology of connected and automated driving vehicles and hence focuses on implementation rather than on pure scientific contribution. Therefore, the paper concentrates on how the modeling of the vehicle, the validation of the model, the hardware-in-the-loop system used in development work, on the hardware architecture and implementation details and presents some experimental results. Controller design details are not presented for the sake of brevity and will be presented in future publications.

The outline of the rest of the paper is as follows. The vehicle modeling is presented in Section 2. The hardware-in-the-loop simulator used in validation and results of the validation study are presented in Section 3. The drive-by-wire architecture of the vehicle is presented in Section 4. The sensors used and the architecture of the vehicle are presented in Section 5. Experimental results of automated driving in the longitudinal and lateral directions are presented in Section 6. The paper ends with conclusions and directions for future work in the last section.

## 2. VEHICLE MODELLING

Simple longitudinal and lateral models of the automated light commercial vehicle presented here are employed for the initial tests of the designed algorithms and controllers. Then, a higher fidelity CarSim vehicle model is used for the final tests before the road experiments. The simple longitudinal and lateral dynamics models used are presented in this section.

### 2.1. Simple Longitudinal and Lateral Automated Vehicle Model

The equations of motion for longitudinal and lateral dynamics of the simple nonlinear automated vehicle model can be expressed as follows:

$$m(a_x - rV_y) = \sum_{i=f,r} F_{xi} \cos\delta_i - F_{yi} \sin\delta_i - (F_{aero} + F_{rr} + F_{hc}) \tag{1}$$

$$m(a_y + rV_x) = \sum_{i=f,r} F_{xi} \sin\delta_i + F_{yi} \cos\delta_i \tag{2}$$

while the equation of motion around the yaw axis is

$$I_z \dot{r} = l_f F_{yf} \cos\delta_f - l_r F_{yr} \cos\delta_r + l_f F_{xf} \sin\delta_f - l_r F_{xr} \sin\delta_r \tag{3}$$

where $F_{xi}$ and $F_{yi}$ are the longitudinal and the lateral tire forces. $f$ and $r$ represent the front and rear tires. $a_x$, $a_y$, $V_x$, $V_y$ and $I_z$ are the longitudinal acceleration at the center of gravity (CG), the lateral acceleration at the CG, the longitudinal velocity at the CG, the lateral velocity at the CG and the moment of inertia about the yaw axis, respectively. Note that for the front wheel steered vehicle considered in this paper $\delta_r = 0$ [1].

The resistive forces in Equation (1) which affect the longitudinal dynamics of the vehicle can be expressed as follows. The aerodynamic drag force $F_{aero}$ is given by

$$F_{aero} = \frac{1}{2} A\rho C_d V^2 \tag{4}$$

where $A$ is the effective frontal area of the vehicle, $\rho$ is the mass density of air, $C_d$ is the drag coefficient, and $V$ is the velocity of the vehicle. The rolling resistance force $F_{rr}$ is determined as

$$F_{rr} = C_{rr} mg \cos(\theta) \tag{5}$$

where $C_{rr}$ is the rolling resistance coefficient and $\theta$ is the road inclination angle. The gravitational slope resistance force $F_{hc}$ is modelled as

$$F_{hc} = mg \sin(\theta). \tag{6}$$

The internal combustion engine (ICE) is modeled using a static engine map that defines the relationship between the inputs of throttle position $\alpha$, the engine speed $\omega$ and the output engine torque $T_{ICE}(\omega, \alpha)$. The engine torque output is transmitted to the wheels through the driveline as torque $T_d$ according to

$$T_d = \eta_t i_t T_{ice}(\omega, \alpha) \tag{7}$$

where $\eta_t$ is a static efficiency factor used to model mechanical losses and $i_t$ is the transmission ratio. These parameters are used to model the transmission of the vehicle.

The moment balance at the center of the wheel is given by

$$I_w \dot{\omega}_i = T_d - T_{bi} - F_{xi} R_w \tag{8}$$

where $I_w$ is the moment of inertia of the wheel, $\omega_i$ is the angular velocity of the $i^{th}$ wheel, $T_{bi}$ is the braking torque on the $i^{th}$ wheel applied through the brake system, $F_{xi}$ is the longitudinal tire force of the $i^{th}$ wheel and $R_w$ is the effective wheel radius.

The longitudinal velocities of the front and rear wheels can be determined as follows:

$$V_{fx} = \sqrt{V_x^2 + (V_y + l_f r)^2} \cos\alpha_f \tag{9}$$

$$V_{rx} = \sqrt{V_x^2 + (V_y - l_r r)^2} \cos\alpha_r \tag{10}$$

where the tire slip angles are

$$\alpha_f = \delta_f - a\tan\left(\tan\beta + \frac{l_f r}{V_x}\right) \tag{11}$$

$$\alpha_r = \delta_r - a\tan\left(\tan\beta + \frac{l_r r}{V_x}\right). \tag{12}$$

The longitudinal wheel slip ratio is defined as

$$s_i = \begin{cases} \dfrac{R_w\omega_i - V_{ix}}{V_{ix}}, & R_w\omega_i < V_{ix} \quad \text{(braking)} \\ \dfrac{R_w\omega_i - V_{ix}}{R_w\omega_i}, & R_w\omega_i > V_{ix} \quad \text{(traction)}, \end{cases} (i = f, r) \tag{13}$$

The Dugoff tire model is used for the calculations of the tire forces as

$$F_{xi} = f_i C_{xi} s_i \tag{14}$$

$$F_{yi} = f_i C_{yi} \alpha_i \tag{15}$$

where $C_{xi}$ and $C_{yi}$ are the longitudinal and the lateral cornering stiffness of the $i^{th}$ wheel. The coefficients $f_i$ are determined using

$$f_i = \begin{cases} 1, & F_{Ri} \leq \dfrac{\mu F_{zi}}{2} \\ \left(2 - \dfrac{\mu F_{zi}}{2 F_{Ri}}\right)\dfrac{\mu F_{zi}}{2 F_{Ri}}, & F_{Ri} > \dfrac{\mu F_{zi}}{2} \end{cases} \tag{16}$$

$$F_{Ri} = \sqrt{(C_{xi} s_i)^2 + (C_{yi} \alpha_i)^2} \tag{17}$$

The simple model given here is implemented in Simulink.

## 2.2. CarSim Vehicle Model

CarSim is a high degrees-of-freedom commercially available vehicle dynamics simulation environment. After the initial simulations in Simulink by using the simple nonlinear vehicle model, a CarSim vehicle model is built and employed for further studies. The subsystems of the vehicle model such as steering, tire, suspension, aerodynamics, powertrain and brake are constructed in CarSim considering the dynamic properties of the automated vehicle. Before the road tests, the control algorithms and also the integration of the automated vehicle subsystems are tested using CarSim simulations. Figure 1 shows the screenshots taken from a CarSim cooperative adaptive cruise control simulation. Also, CarSim/Simulink block diagrams can be seen in Figure 1.

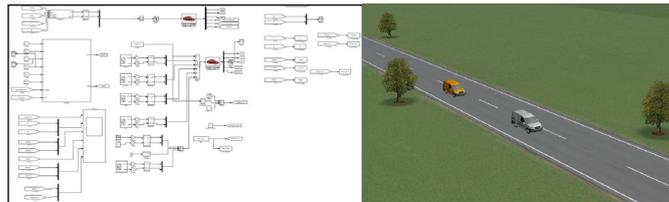

**Figure 1.** CarSim/Simulink block diagrams and animation screen

In addition to offline simulations, real time hardware-in-the-loop (HiL) simulations are conducted to analyze the integration of the designed control systems, sensor interfacing algorithms and hardware. Figure 2 shows the hardware-in-the-loop system used in this study. It consists of a PC, a dSPACE Ecoline simulator, an Electronic Stability Control (ESC) Electronic Control Unit (ECU) module, a valve signal measurement unit, steering wheel angle sensor, sensor cluster consisting of yaw rate and longitudinal and lateral acceleration sensors, vehicle display panel, body computer and fuse box. Further details can be found in [6]. While this HiL simulator is originally built for testing ESC algorithms, it is also well suited for testing automated steering control. The dSPACE MicroAutoBox control unit, which is used for low level actuator control in the automated vehicle, is added to this HiL setup for the automated steering HiL simulations. The addition of the automated brake test rig shown in Figure 3 enables the testing of automated braking control also, in this HiL simulation test system.

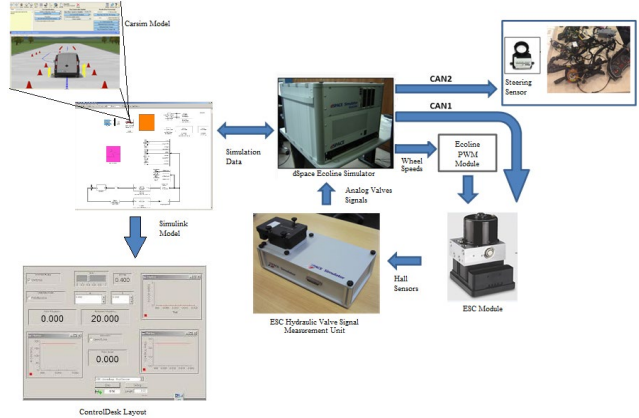

**Figure 2.** Hardware-in-the-loop simulation system

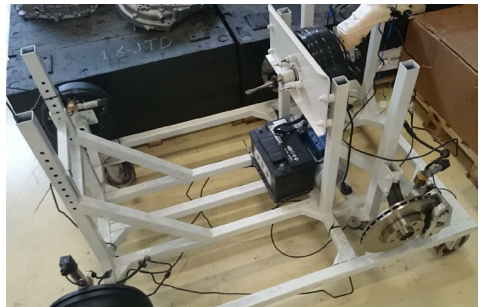

**Figure 3.** Automated braking hardware in the HiL system

## 3. MODEL VALIDATION RESULTS

In order to validate the vehicle dynamics responses of the CarSim model, vehicle road tests are carried out using an instrumented test vehicle. The standard tests like constant radius turning, ramp steering, double lane change, sine-with-dwell are conducted. In CarSim, the same experimental environment is built and the same velocity and the steering inputs are provided to the CarSim vehicle model. The experimentally obtained results are compared with CarSim vehicle model responses. According to the results, CarSim vehicle model parameters such as Pacejka tire values, CoG height and the roll stiffness of the vehicle are tuned to obtain good matching between experimental results and simulation results. The simulations are performed in real-time using the HiL simulator shown in Figure 2. Out of the many tests that are carried out, the standard ISO 3888-1 double lane change results are given in this paper. In the double lane change test, the vehicle tries to change its lane and then go back to its initial lane as shown in Figure 4.

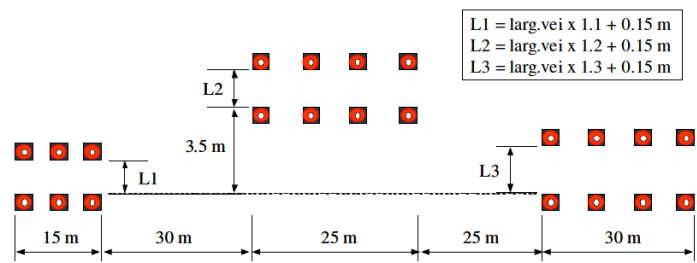

**Figure 4.** ISO 3888-1 Double lane change test [7]

Figure 5 and Figure 6 show the steering wheel angle and the vehicle velocity signals of the vehicle obtained during the double lane change test. These signals are used in CarSim (labeled as HiL test in the figures) as input signals and the simulations are performed.

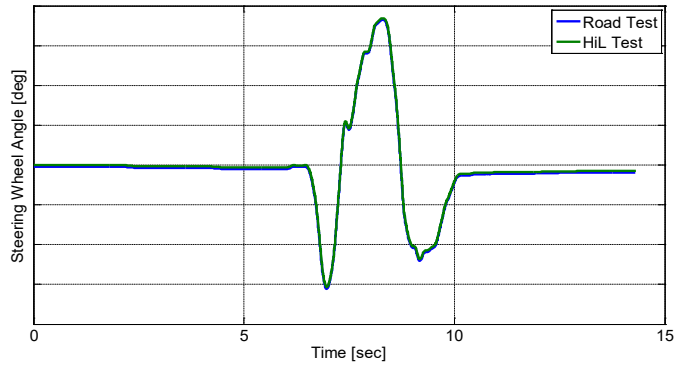

**Figure 5.** Steering wheel input for the double lane change test

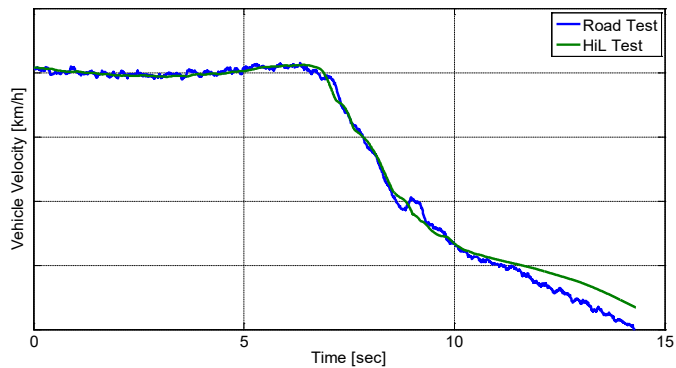

**Figure 6.** Vehicle velocity for double lane change test

Figure 7 and Figure 8 show the vehicle yaw rate and lateral acceleration results obtained from both experiment and HiL simulation. The results indicate that the validated vehicle model and the experimental vehicle responses match each other quite well. It is concluded that the validated Carsim model can be used to describe the steady state dynamics of the experimental vehicle in the simulations.

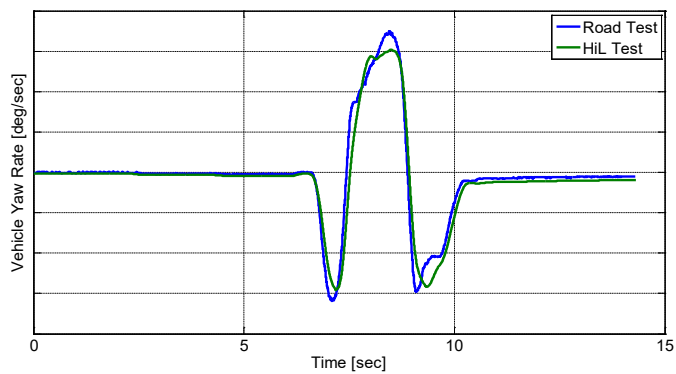

**Figure 7.** Vehicle yaw rate for the double lane change test

## 4. DRIVE-BY-WIRE SYSTEMS

In order to control the throttle, braking and steering actuators, a dSPACE MicroAutoBox (MABX) electronic control unit is used. Automation of throttle, brake and steering are presented in this section of the paper.

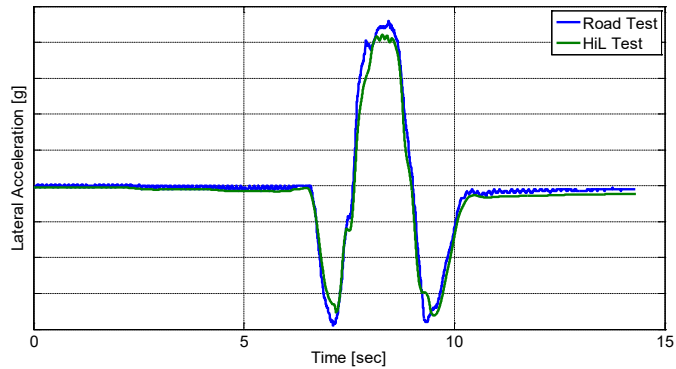
**Figure 8.** Lateral acceleration for the double lane change test

### 4.1. Throttle and Brake Control

The upper and lower level controllers for longitudinal direction control of the vehicle use throttle and brake actuation. Depending on the chosen automated driving function such as cruise control (CC), adaptive cruise control (ACC) or cooperative adaptive cruise control (CACC), the longitudinal direction controller (PI type for CC and PD type for ACC/CACC) determines the desired acceleration value, $a_{desired}$.

For positive $a_{desired}$ values, the related lower level controller produces the appropriate throttle command using the inverse-engine torque map. The throttle command is then converted to an analog signal in the MABX controller and is fed to the interface electronics that changes the potentiometer reading of the accelerator pedal which is fed to the internal combustion engine controller. If the driver pushes on the accelerator pedal, control authority is switched over from the automated driving system to the driver.

For negative $a_{desired}$ values, the related lower level controller determines an appropriate duty cycle (%) of the PWM signal according to the relation between braking deceleration and duty cycle to control the active vacuum booster used for automated braking. This vacuum booster can be controlled manually (driver) or electrically (automated braking). Due to the high current requirement for the servo system of the active vacuum booster, a relay circuit board is implemented as a driver.

### 4.2. Steering Control

In order to control the automated steering, a SMART motor by Moog Animatic is integrated into the light commercial vehicle. The main characteristics of the SMART motor are that it can be programmed, its drive electronics are integrated into the motor and it has a CAN interface along with the customary analog interface for command signals.

A PI type position controller is implemented in the dSPACE MABX electronic control unit for steering control. After the calculation of the desired angle by the "Path Follower" controller, this PI position controller generates an analog signal to control the SMART motor. A program created in the SMART motor interprets this signal and produces a PWM signal to drive the motor. Instead of the SMART motor`s individual encoder, the vehicle`s own steering angle sensor (SAS) is used for better position feedback. The steering control system schematic is shown in Figure 9.

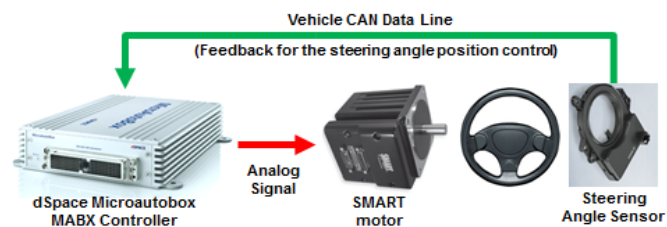
**Figure 9.** Steering control schematic system

### 4.3. Path Following Controller for Autonomous Driving

For demonstration of autonomous driving, a simple path following control algorithm is implemented as shown in Figure 10. Future work will use the algorithm of the authors in [4] which handles both position and orientation errors, passes smooth curves through the GPS way points and results in a much smoother trajectory with smaller path tracking errors.

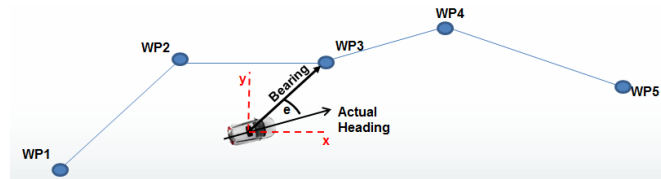

**Figure 10.** Path following algorithm

The desired path is created by choosing the desired GPS waypoint positions (WPs) using a map application (Google Maps etc.). After sending the WPs to the controller, the path following algorithm creates a bearing vector between the vehicle position and the target WP. Then, the angle between the bearing vector and the heading vector is calculated and a PID type controller, shown in Figure 11, keeps the angle at zero value while controlling the steering.

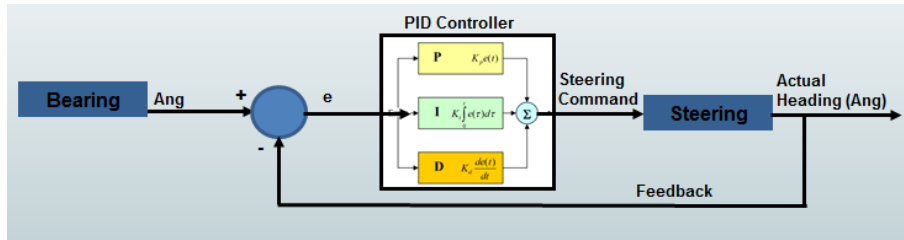

**Figure 11.** PID type path following algorithm

## 5. ENVIRONMENTAL SENSING AND OVERALL ARCHITECTURE

### 5.1. Environmental Sensing

A Delphi ESR Radar, an Ibeo LUX four channel lidar and an Xsens GPS sensor with built in IMU and INS algorithm are used for the environmental sensing and localization. The radar sensor is used to detect the range and the speed of the followed vehicle in the cooperative/adaptive control (CACC and ACC) system. The lidar sensor is mostly used for detecting closer objects in autonomous driving.

The position of the vehicle is determined by the Xsens GPS sensor. Even though it does not provide higher precision (in centimeter level) position values, it has been chosen because of its low cost and very good repeatability performance.

The GPS signal errors (about 1-2 meter) are compensated for by the lidar sensor during low speed automated driving. If a sideway or an object in the direct path is detected, the path following controller generates low speed steering correction angles as shown in Figure 12. The more recent work of the authors uses SLAM or map building and map matching algorithms using a 3D lidar for high accuracy localization which is not treated in this paper.

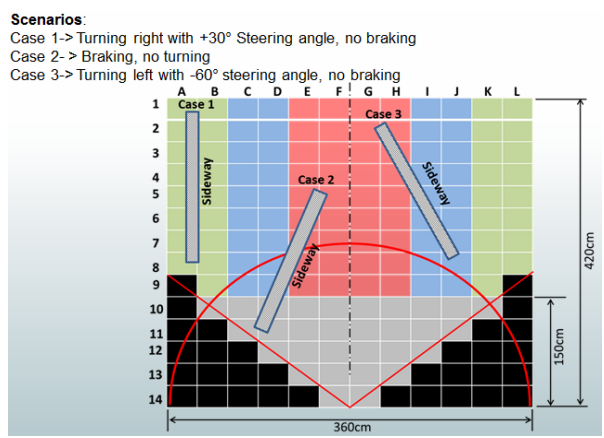

**Figure 12.** Steering correction scenarios

To overcome the magnetic interferences and improve the heading information, an additional digital compass, HMC5883L, is integrated into the system. The average of the digital compass's heading value and the Xsens GPS sensor's heading value is used in the path following controller.

For the cooperative adaptive controller (CACC) system, the vehicle is equipped with a modem using IEEE 802.11p protocol. The modem receives the followed vehicle's acceleration, GPS latitude and longitude values from the followed vehicle's sender modem. The received acceleration value is feed-forwarded to the current adaptive cruise controller (ACC) to establish a cooperative adaptive cruise control (CACC) system as in [1].

## 5.2. Overall Architecture

The installed actuators, sensors, controllers and modem are shown in the overall architecture diagram of Figure 13.

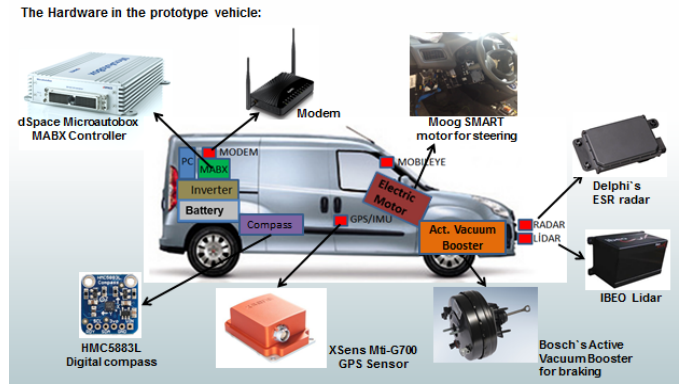

**Figure 13.** Actuators, sensors, controlers and modem in the vehicle

As shown in Figure 14, the modem data (used for CACC) and lidar data are sent to the PC via an Ethernet switch. The received data are then combined with the GPS sensor's latitude, longitude, and heading data received via the USB port, and the heading data received from the digital compass via I2C communication line. After assigning an identifier number, each data is then converted to 8 byte long UDP (user datagram protocol) data packages and sent to the UDP to CAN converter board.

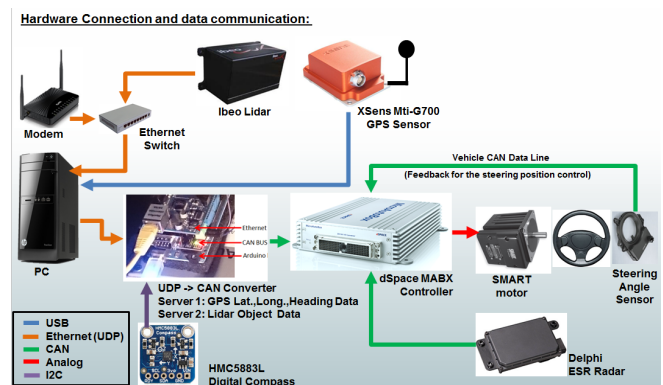

**Figure 14.** Data communication in vehicle

The UDP to CAN data converter board is developed using an Arduino microcontroller, an ethernet shield (to receive the UDP data packages) and a CAN bus shield (to convert the UDP packages to CAN signals and send to the MABX controller via vehicle's CAN data line). The created board is shown in Figure 15. As shown in Figure 14, the radar data is directly sent to the MABX controller via the vehicle's CAN data line. In addition, the steering angle value is read from the steering angle sensor (SAS) via the CAN data line for the steering angle position control.

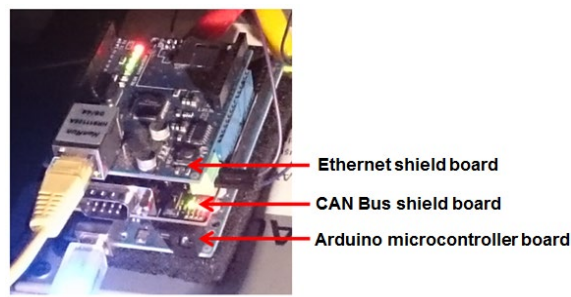

**Figure 15.** Data communication in vehicle

## 6. EXPERIMENTAL RESULTS

Example experimental results are presented in this section to demonstrate the working of the automated driving light commercial vehicle presented in this paper. The example experimental results include a CACC test in the longitudinal direction and path following in the lateral direction, presented separately. The CACC algorithm similar to that in [1] is first tested on the HiL system as shown in Figure 1. The ACC part of the algorithm uses the radar and the CACC part uses the IEEE 802.11p V2V modem and acceleration sensor and GPS sensors. The preceding vehicle is driven by a driver while the ego vehicle which is the automated driving vehicle of this paper follows autonomously.

The velocity difference between the preceding and ego vehicles during one of the tests is shown in Figure 16. In the path following experiment, a cruise controller is used in the longitudinal direction as there is only one vehicle. The simple automated steering and collision avoidance algorithms presented earlier in this paper are implemented in the lateral direction for steering control. The GPS waypoints are obtained automatically using an interface program. The Simulink block diagram used in the implementation is shown in Figure 17. The results of an example path following experiment are shown in Figure 18. The red arrows indicate the direction of travel of the vehicle. The automated steering algorithm of the authors in [8] will be used in future implementation of automated steering.

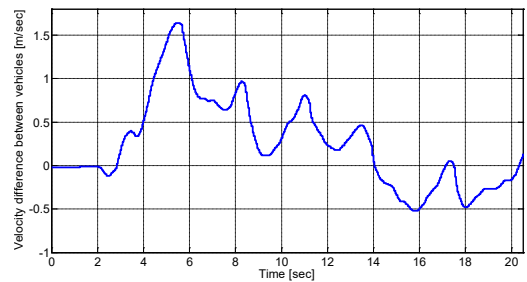

**Figure 16.** Velocity difference between preceding and ego vehicles during CACC testing

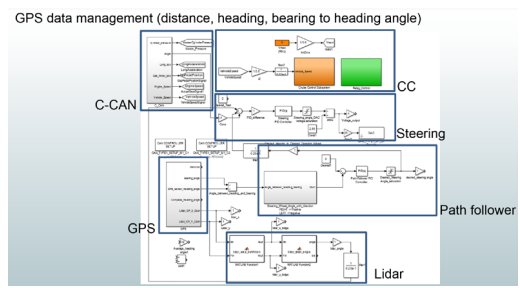

**Figure 17.** Simulink block diagram of implemented steering controller

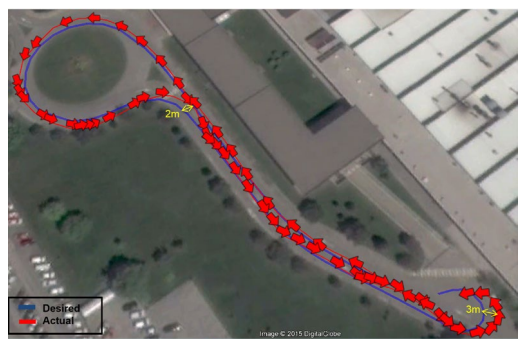

**Figure 18.** Results of automated steering experiment

It is worthwhile to cite some other relevant references and ones that can be used in future work here. These include research on active safety and ADAS systems like those in references [9–13] which have been followed by research on robust and energy saving methods like those in references [14,15] and more recent research on autonomous driving like those in references [16–22]. Research on traffic flow improvement and energy consumption reduction can be found in references like [15–19]. Research on safety and other research that can be used in future work can be found in [17,28–38].

# 7. CONCLUSIONS

This paper presented the software, hardware, control and decision making architecture of a light commercial vehicle that was modified into an automated driving vehicle. Simple Simulink and more detailed CarSim vehicle dynamic models were introduced and validated using experimental responses of the vehicle obtained in a test track. A HiL setup used for preliminary evaluation and validation of the results was illustrated. The actuators used for by-wire automation of the vehicle were presented. The localization and environmental sensors used were explained in detail. The paper ended with exemplary test results with CACC testing in the longitudinal direction and automated path following steering control in the lateral direction.

# ACKNOWLEDGEMENTS

The authors from TOFAŞ acknowledge the support of TÜBİTAK, the Scientific and Technological Research Council of Turkey through its grant TEYDEB - 1120238. The other authors acknowledge the support of TOFAŞ. The authors thank Furkan Gökmen of TOFAŞ for his help with the vehicle and the HiL setup.